\documentclass{article}
\usepackage{RF}

\usepackage{times}
\usepackage{xcolor}
\usepackage{soul}
\usepackage[utf8]{inputenc}
\usepackage[small]{caption}

\usepackage{graphicx}
\usepackage{amsfonts}
\usepackage{amsmath}
\usepackage{url}
\usepackage{booktabs}

\usepackage{enumerate}
\title{Load Balanced GANs for Multi-view Face Image Synthesis}

\author
{
	Jie Cao\textsuperscript{1,2,4}, Yibo Hu\textsuperscript{1,2,4}, Bing Yu\textsuperscript{5}, Ran He\textsuperscript{1,2,3,4} and Zhenan Sun\textsuperscript{1,2,3,4}\\
	\textsuperscript{1}National Laboratory of Pattern Recognition, CASIA \\
	\textsuperscript{2}Center for Research on Intelligent Perception and Computing, CASIA \\
	\textsuperscript{3}Center for Excellence in Brain Science and Intelligence Technology, CAS \\
	\textsuperscript{4}University of Chinese Academy of Sciences, Beijing, 100049, China \\
	\textsuperscript{5}Noah's Ark Lab of Huawei Technologies \\
	{\tt\small \{jie.cao,yibo.hu\}@cripac.ia.ac.cn, yubing5@huawei.com, \{rhe, znsun\}@nlpr.ia.ac.cn} \\
}
\begin{document}

\maketitle

\begin{abstract}
Multi-view face synthesis from a single image is an ill-posed problem and often suffers from serious appearance distortion. Producing photo-realistic and identity preserving multi-view results is still a not well defined synthesis problem. This paper proposes Load Balanced Generative Adversarial Networks (LB-GAN) to precisely rotate the yaw angle of an input face image to any specified angle. LB-GAN decomposes the challenging synthesis problem into two well constrained subtasks that correspond to a face normalizer and a face editor respectively. The normalizer first frontalizes an input image, and then the editor rotates the frontalized image to a desired pose guided by a remote code. In order to generate photo-realistic local details, the normalizer and the editor are trained in a two-stage manner and regulated by a conditional self-cycle loss and an attention based L2 loss. Exhaustive experiments on controlled and uncontrolled environments demonstrate that the proposed method not only improves the visual realism of multi-view synthetic images, but also preserves identity information well.
\end{abstract}

\section{Introduction}
Multi-view face image synthesis has plenty of applications in various domains including pose-invariant face recognition, virtual and augmented reality, and computer graphics. Although humans can easily conceive different views of a face in mind when seeing it, making the computer have this conceive (synthesis) ability is an appealing and long-standing challenge. Traditional methods resort to 3D Morphable Model (3DMM) \cite{blanz1999morphable} to address this challenge. They build 3D face model as reference and then synthesize face images with new angles through model fitting. Although these 3D methods can synthesize or rotate a face image to some extent, their synthesis results are often not photo-realistic.

Recently, face synthesis models based on convolutional neural networks (CNNs) have drawn much attentions. These methods are built on black-box models and do not depend on 3D facial shape. Without explicitly modeling a face, they produce the output under the control of remote code \cite{ghodrati2015towards,yim2015rotating}. For instance, if the remote code of yaw angle is set to ${30^\circ }$, then the networks will automatically rotate an input image with an arbitrary pose to ${30^\circ }$. Recently, with the application of Generative Adversarial Networks (GAN) in multi-view face image synthesis, much progress has been made \cite{DRGAN,zhao2017dual}. However, when desired synthesis pose tends to be larger, it is still easy for humans to distinguish the synthesized images from the genuine ones. 

\begin{figure}[!tbp]
\begin{center}
\includegraphics[width=.45\textwidth]{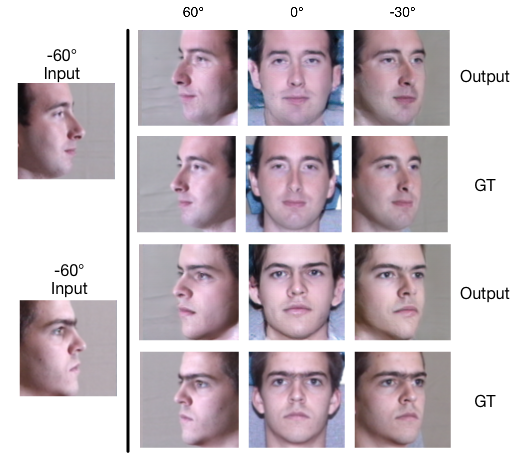}
\end{center}
\caption{Face rotation results by our LB-GAN. According to the degrees on the top for each column on the right-hand, the inputs are rotated to a specified pose. GT stands for the ground truth.}
\label{show}
\end{figure}

This paper addresses the challenging multi-view face image synthesis problem by simplifying it into two subtasks, resulting in a new method named Load Balance Generative Adversarial Networks (LB-GAN). Concretely, we employ two pairs of GAN whose generators cooperate with each other. The first GAN consists of a generator termed as face normalizer and the corresponding discriminator. The face normalizer only focuses on frontalizing face images. The second GAN consists of another generator termed as face editor and its discriminator. The face editor takes frontal view face images as additional inputs and rotates the input image to a specified pose according to a given remote code. We combine the normalizer and the editor together to rotate the yaw angle of an input face to any specified one. Some synthesized samples by LB-GAN are shown in Fig. \ref{show}.

We employ a two-stage strategy to train LB-GAN. In the first stage, only the face normalizer and its discriminator are trained through the conventional manner for GAN \cite{GAN}. After plausible results have been produced by the normalizer, we begin training the whole model in the second stage. Considering that noisy backgrounds of face images captured in unconstrained environments will degrade visual realism of the result severely, we propose a novel conditional self-cycle loss and an attention based $L2$ loss to tackle this problem. Experimental results on Multi-PIE and IJB-A show that our method can produce photo-realistic multi-view face images. Besides, the performance of pose-invariant face recognition is boosted through our synthetic results. In summary, the main contributions of our work are: 

\begin{enumerate}[1)] 
\item We propose LB-GAN that simplifies the ill-posed multi-view face synthesis problem into two well constrained ones.
\item Trained in a novel two-stage method, our model can preserve abundant identity information while rotating a face to arbitrary poses.
\item Profiting from the conditional self-cycle loss and attention based $L2$ loss, our model is robust to noisy environments.
\item Experimental results show that our model produces photo-realistic multi-view face images and obtains state-of-the-art cross-view face recognition performance under both controlled and uncontrolled environments.
\end{enumerate}

\section{Related Work}

\subsection{Face Frontalization}
Face frontalization can be regarded as a single view face image synthesis problem, i.e., producing the frontal face images is the only consideration. To eliminate the influence of poses in face recognition or other facial analysis tasks, face frontalization has been widely studied in recent years. 3D-based models \cite{dovgard2004statistical,hassner2013viewing,hassner2015effective,zhu2015high,ferrari2016effective} are proposed for frontalization in controlled environments. Besides, deep learning models are also very competitive, e.g., CNNs \cite{zhu2013deep,zhu2014multi}, auto-encoders \cite{zhang2013random,kan2014stacked} and recurrent neural networks (RNNs) \cite{yang2015weakly}. At present state-of-the-art face frontalization methods in controlled \cite{huang2017beyond} and in-the-wild \cite{zhao2017dual} settings are both based on GAN. 

For face frontalization and other face synthesis tasks, the capacity of identity preservation is mainly evaluated through face recognition. To this end, \cite{DRGAN} extract pose-robust identity representations from the face generator for recognition, while \cite{hassner2015effective,huang2017beyond} directly use the synthesized face images for recognition.

\subsection{Generative Adversarial Networks}
GAN is a novel deep framework proposed by \cite{GAN}. GAN can be regarded as a two-player non-cooperative game model. The main components of GAN, generator and discriminator, are rivals of each other. The generator tries to map some noise distribution to the data distribution. The discriminator tries to distinguish the fake data produced by the generator from the real data. In practice, the parameters of the generator and the discriminator are trained alternately until convergence. The most significant contributions of GAN is the remarkable improvement on visual realism. Conditional GAN is proposed by \cite{mirza2014conditional} to send conditional information to both the generator and discriminator. To deal with unpaired data, \cite{zhu2017unpaired} propose CycleGAN.

\section{Proposed Methods}
%-------------------------------------------------------------------------

\begin{figure*}[!tbp]
\centering 
\includegraphics[width=\textwidth]
{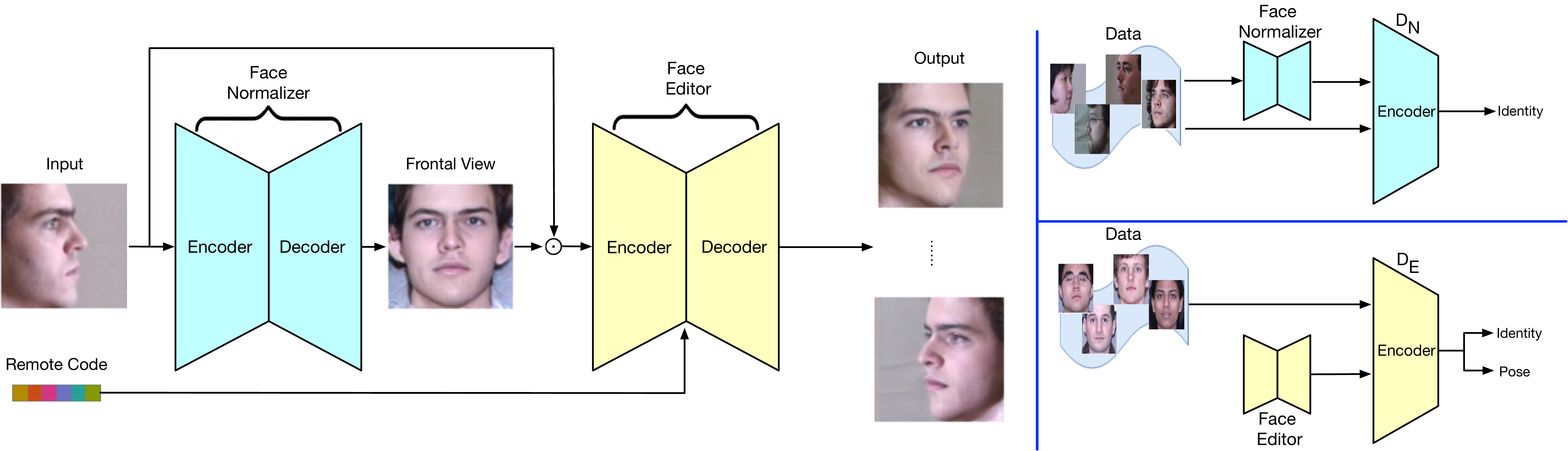}
\caption{The framework of our LB-GAN for multi-view face image synthesis.}
\label{LB-GAN}
\end{figure*}

Assume there are $n_{id}$ subjects in the training set and each face image $\mathbf{x}$ has corresponding identity label $y^{id}$ and pose label $y^p$. $y^{id} \in\{1, 2, \cdots, n_{id}-1, n_{id}\}$. $y^{p}\in\{-90^{\circ}, -75^{\circ}, \cdots 0^{\circ}, \cdots 90^{\circ}\}$. So $y^p$ has $n_p=13$ discrete possible values. Remote code $\mathbf{c}$ is a $n_p$-dimensional one hot vector. We assign the $c^*$th element in $\mathbf{c}$ to $1$ only if we want to change the pose of input to the $c^*$th type. Our goal is to train a model which takes a remote code $\mathbf{c}$ and maps the given $\mathbf{x}$ to a new face image $\hat{\mathbf{x}}$. $\hat{\mathbf{x}}$ should meet the following three requirements: (1) the visualization of $\hat{\mathbf{x}}$ is realistic, (2) the identity of $\hat{\mathbf{x}}$ remains the same as $\mathbf{x}$, (3) the pose is altered according to the specified $\mathbf{c}$.

\subsection{Model Structure}\label{S2}

As illustrated in Fig. \ref{LB-GAN}, our proposed LB-GAN consists of a pair of GAN to address the multi-view face synthesis problem. The first GAN is composed of the face normalizer $G_N$ and the corresponding discriminator $D_N$. Similarly, the second GAN is composed of the face editor $G_E$ and $D_E$. During the test phase, $G_N$ first takes $\mathbf{x}$ and transforms it into frontal view face image (we denote the pose label of frontal view face images as $y^{p^*}$ below), then $G_E$ takes $\mathbf{x}$, the output of $G_N$ and the remote code $\mathbf{c}$ to produce the desired $\hat{\mathbf{x}}$.

In the training stage, $D_N$ takes fake images produced by $G_N$ or genuine images draw from datasets with pose label $y^{p^*}$. Similar with \cite{DRGAN}, the goal of $D_N$ is giving explicit identities of input images rather than simply judging whether they are produced by $G_N$. $D_N(\mathbf{x})$ is the prediction for the identity made by $D_N$. $D_N(\mathbf{x})=[D^{1}_{N}(\mathbf{x}),D^{2}_{N}(\mathbf{x}),\cdots, D^{n_{id}}_{N}(\mathbf{x}),D^{y^{id^{*}}}_{N}(\mathbf{x})]$, where $D^{i}_{N}(\mathbf{x})$ stands for the probability that the identity label of $\mathbf{x}$ equals to $i$. The identity labels of produced images are all $y^{id^{*}}$. Fed by $\mathbf{x}$ with $y^{id}$, $G_N$ aims to fool $D_N$ into believing the produced image having identity label $y^{id}$. The objective functions of $D_N$ and $G_N$ can be formulated as:

\begin{equation}
\label{D1}
\begin{split}
\mathop {\max }\limits_{\Theta_{D_N}} {V}({\Theta_{D_N}}) &= {\mathbb{E}_{\substack{\mathbf{x},{y^{id}}\sim{p_{m}}}}}[\log {D_{N}^{{y^{id}}}}(\mathbf{x})] \\
&+ {\mathbb{E}_{\substack{\mathbf{x},{y^p},{y^{id}}\\\sim{p_{data}}}}}[\log {D_{N}^{{y^{id^{*}}}}}(G_{N}(\mathbf{x})]
\end{split}
\end{equation}

\begin{equation}
\label{G1}
\mathop {\max }\limits_{\Theta_{G_N}} {V}({\Theta_{G_N}}) = {\mathbb{E}_{\substack{\mathbf{x},{y^p},{y^{id}}\\\sim{p_{data}}}}}[\log {D_{N}^{{y^{id}}}}(G_{N}(\mathbf{x}))]
\end{equation}

\noindent where $\Theta_{D_N}$ and $\Theta_{G_N}$ denote the parameter sets of $D_N$ and $G_N$ respectively, ${p_{data}}={p_{\mathbf{x},{y^p},{y^{id}}}}(\mathbf{x},{y^p},{y^{id}})$, ${p_{m}} = {p_{\mathbf{x},{y^{id}}|{y^p} = {{y}^{{p}^*}}}}(\mathbf{x},{y^{id}})$. The first item in Eq. (\ref{D1}) pushes $D_N$ to recognize the identities of the subjects in the training set, and the second one pushes $D_N$ to find the images produced by $G_N$. In the meantime, to maximize Eq. (\ref{G1}), $G_N$ has to keep the identity information of the input well preserved. Although $D_N$ does not directly discriminates the pose label, $G_N$ has to transform the pose of its input into $y^{p^*}$ to be in accordance with genuine data. In such an adversarial training procedure, $D_N$ will be able to distinguish the real and the fake, and $G_N$ will be able to produce photo-realistic images.

$D_E$, which can be regarded as a multi-task classifier, predicts poses as well as identities. The predictions of the identity and the pose made by $D_E$ are denoted as $D_E(\mathbf{x})$ and $D_{E_p}(\mathbf{x})$ respectively. Their definitions are similar with $D_N(\mathbf{x})$. Guided by the remote code $\mathbf{c}$, the goal of $G_E$ is to alter the pose of input to the $c^{*}$th type without being discovered by $D_E$. $\mathbf{c}$ is added to $G_E$ through the way proposed by \cite{salimans2016improved}. Formally, $G_E$ and $D_E$ are optimized as follows: 

\begin{equation}
\label{D2}
\begin{aligned}
\mathop {\max }\limits_{\Theta_{D_E}} {V}({\Theta_{D_E}}) &= {\mathbb{E}_{\substack{\mathbf{x},{y^p},{y^{id}}\\\sim{p_{data}}}}}[\log D_{E}^{{y^{id}}}(\mathbf{x}) + \log {D_{E_p}^{{{y^p}}}(\mathbf{x})} \\ 
&+ \log D_{E}^{{y^{id^{*}}}}({G_E}(\mathbf{x},G_N(\mathbf{x}),\mathbf{c})]
\end{aligned}
\end{equation}

\begin{equation}
\label{G2}
\begin{aligned}
\mathop {\max }\limits_{\Theta_{G_E}} {V}(\Theta_{G_E}) &= {\mathbb{E}_{\substack{\mathbf{x},{y^p},{y^{id}}\\\sim{p_{data}}}}}[\log {D_{E_p}^{{{c^*}}}(G(\mathbf{x},G_N(\mathbf{x}),\mathbf{c}))} \\
&+ \log D_{E}^{{y^{id}}}({G_E}(\mathbf{x},G_N(\mathbf{x}),\mathbf{c}))]
\end{aligned}
\end{equation}

\noindent where $\Theta_{D_E}$ and $\Theta_{G_E}$ denote the parameter sets of $D_E$ and $G_E$ respectively. Since $D_E$ is trained to discriminate poses as well, $G_E$ need to change the pose of its input but keep the visual realism and the identity information well-preserved. Note that $G_E$ takes both $\mathbf{x}$ and the face frontalized by $G_N$ as input. Different from previous approaches that synthesize multi-view face images by a single generator, our $G_E$ get more information from $G_N$ to obtain robustness of dealing with variant poses. The frontalized face will be helpful for rotating face with extreme poses, and the original input face will contribute more for identical and symmetric conditions (e.g., rotate $60^{\circ}$ to $60^{\circ}$ and rotate $-30^{\circ}$ to $30^{\circ}$).

\subsection{Two-stage Training Method}\label{S3}

The training process of our LB-GAN is two-staged. In the first stage, we only train the face normalizer and its discriminator in the alternative and adversarial manner \cite{GAN}. We stop the process when visually appealing results have been generated by $G_N$. Then in the second stage, we train the whole model. We find making the parameters of $G_N$ near-optimal first will stabilize the second training procedure and guarantee better final performance. Specifically, we optimize the parameters by Adam optimizer \cite{kingma2014adam} with a learning rate of 2e-4 and momentum of 0.5. The first training stage lasts for 20,000 iterations. In the second stage, the learning rate for $G_N$ and $D_N$ is reduced to a quarter. We train 4 iterations for optimizing $G_N$ and $G_E$, and then 1 iteration for $D_N$ and $D_E$. The batch size is set to 24. Note that extra regularization items, which will be discussed in section \ref{S4}, are added in the second stage.

\subsection{Regularization Items}\label{S4}

\textbf{Attention based $\mathbf{L2}$ Loss.} $L2$ loss is a common choice for measuring the difference of two images, and every pixel is treated equally. However, to make synthesized face images realistic and characteristic, some key facial parts should be emphasized, like eyes, mouth, nose, etc. Further, for those images captured in real life condition, the styles of clothes and hair of subjects and the background tend to change frequently. So minimizing $L2$ loss will make those regions blurry. To this end, attention based $L2$ loss denoted as $L{2^{\prime}}$ is proposed:

\begin{equation}
L{2^{\prime}}(\mathbf{x,\hat{x}}) = {\left\| {\mathbf{(x - \hat{x})} \circ M} \right\|_2}
\end{equation}

\noindent where the operator $\circ$ denotes the Hadamard product, and $M$ is a mask whose entries are set to $1$ for the region of interest and $0$ otherwise. Through adding $M$, our model is guided to concentrate on synthesizing convincing facial images and avoid putting too much attention on unnecessary details in the background.

Obviously, the optimal $M$ should exactly cover the facial part of $\mathbf{x}$ and exclude the other parts. But choosing the optimal $M$ for calculating $L{2^{\prime}}$ loss is very expensive. Laborious manual annotation or face parsing algorithm with high accuracy is required. We sidestep this demand by loosening the restriction of $M$. Concretely, $M$ only covers a few image patches that contain key parts of face. The location of the patches can be determined by the landmarks, which are also used for face image preprocessing. Since the input face image will be scaled and aligned, the proper patch sizes and locations for one image also works well for the others, we keep the the patch sizes and locations fixed for all images. The specify choice of $L{2^{\prime}}$ will be given in the experiment section.

\textbf{Conditional self-cycle loss ($L_{csc}$).} $L_{csc}$ is come up with such an observation: if $\mathbf{c}$ exactly matches the pose label $y^{p}$, e.g., the input $\mathbf{x}$ is a frontal view image and the remote code $\mathbf{c}$ sets the output to be frontal view as well, then $\mathbf{x}$ itself is the optimal result, so $\hat{\mathbf{x}}$ should be the same with $\mathbf{x}$ in such a case (the case is denoted as ``$\mathbf{x}$ is the optimal output below''). Our $L_{csc}$ is formulated as:

\begin{equation}
L_{\csc } =
\begin{cases}
 & {L_2^{\prime}}(\mathbf{x}, \hat{\mathbf{x}}),~if~\text{$\mathbf{x}$ is the optimal output}\\
 & \quad 0, \quad \quad \quad otherwise
\end{cases}
\end{equation}

We use $L{2^{\prime}}$ to measure the difference between $\mathbf{x}$ and $\hat{\mathbf{x}}$. The name of $L_{csc}$ is similar with cycle consistency loss ($L_{cyc}$) proposed by \cite{zhu2017unpaired}. However, $L_{cyc}$ is designed to enforce the networks to find the corresponding relationship when training with unpaired data and is originally proposed for domain transfer problem, such as image-to-image translation. In contrast, $L_{csc}$ enforces the networks to avoid redundant operation on the input, i.e., the networks are encouraged to keep the input identical in some circumstance. Besides, the input and the output are in the same semantic domain in the face image synthesis problem. 

\section{Experiments and Analysis}
\subsection{Experimental Settings}

\indent\indent\textbf{Datasets.} Three datasets are involved in our experiments: Multi-PIE \cite{gross2010multi}, IJB-A \cite{Klare2015Pushing} and CASIA-WebFace \cite{yi2014learning}. Multi-PIE is established for studying on PIE (pose, illumination and expression) invariant face recognition. 20 illumination conditions, 13 poses within $\pm {90^{\circ}}$ yaw angles and 6 expressions of 337 subjects were captured in controlled environments. IJB-A is another database with large pose variations. It has 5, 396 images and 20, 412 video frames of 500 subjects. CASIA-WebFace is a large-scale dataset containing 10,575 subjects and 494,414 images. The collection process started from the well-structured information in IMDB and then continued by using web crawler.

For experiments on Multi-PIE, we use the first 200 subjects for training and the rest 137 for testing. Each testing identity has one gallery image from his/her first appearance. Hence, there are 72,000 and 137 images in the probe and gallery sets respectively. There are no overlap subjects between the training and testing sets. To test the performance on IJB-A, we train our model on Multi-PIE and CASIA-WebFace. We follow the testing protocol in \cite{DRGAN}.

\textbf{Data Preprocessing.} Face images in those datasets are normalized to $96\times 96$ before fed into our model. Image intensities are linearly scaled to the range of $[-1,1]$. We use the landmarks of the centers of eyes and mouth to normalize images by the method proposed by \cite{he2017learning}. Note that the normalization in this step is apparently different from the function of $G_N$. The mask $M$ consists of three parts: two $15 \times 15$ patches centered at the eyes and a $20 \times 20$ patch centered at the mouth center. 

\textbf{Implementation Details.} We employ the improved version \cite{DRGAN} of CASIA-Net \cite{yi2014learning} to see if our novel structure can push the limit set by them. The CASIA-Net, which can be regarded as an encoder, is able to transform an input image into an identity representation. To map the representation back to image, we build the decoder through replacing the convolution layer with transposed convolution layers. $G_N$ and $G_E$ have the encoder-decoder structure, $D_N$ and $D_E$ only have the encoder structure. Fully connected layers are added for discriminators to predict the identities and the poses. The remote code $\mathbf{c}$ is injected into the bottleneck layer of $G_E$.

\subsection{Comparison Results}

\begin{figure*}[!tbp]
\begin{center}
\includegraphics[width=\textwidth]{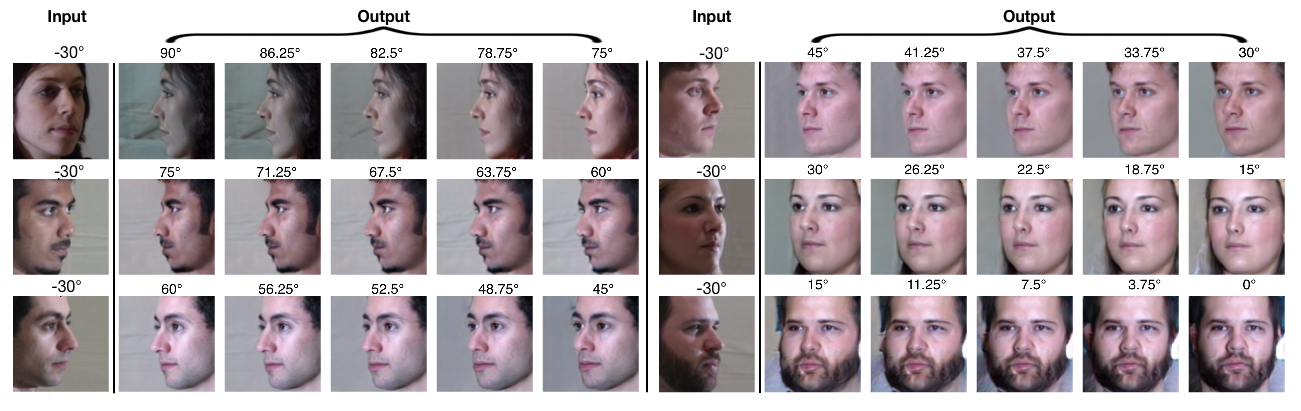}
\end{center}
\caption{Face rotation results on Multi-PIE. Each input is rotated to the specified yaw angle which is indicated by the degree above the output.}
\label{pose}
\end{figure*}

\begin{figure*}[!tbp]
\begin{center}
\includegraphics[width=\textwidth]{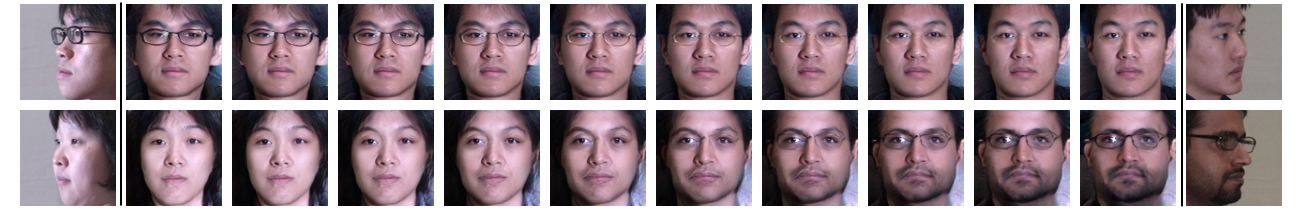}
\end{center}
\caption{Synthesizing new samples through interpolating identity representations. For each row, the samples in the middle are synthesized by the interpolation of the identity representations of the far left and the far right faces.}
\label{ID}
\end{figure*}

To demonstrate the effectiveness of our method, we make comparisons with several state-of-the-art ones. The performances are evaluated both qualitatively and quantitatively. Specifically, three aspects are considered: the visual quality, the performances on pose-invariant face recognition and head pose estimation.

\begin{table}[!tbp]
\small
\centering
\caption{Benchmark comparison of identification rate (\%) across poses on Multi-PIE. Methods marked with $\dag$ can only produce frontal view face images. Methods marked with $*$ are ordinary face recognition ones that are not designed for pose invariant recognition.}
\label{multipie}
\scalebox{0.9}
{
\begin{tabular}{ccccccc}
\toprule
Method& $\pm 15^{\circ}$& $\pm 30^{\circ}$ & $\pm 45^{\circ}$ & $\pm 60^{\circ}$ & $\pm 75^{\circ}$ & $\pm 90^{\circ}$\\
\midrule
DR-GAN & 94.9 & 91.1 & 87.2 & 84.6 & -     & -   \\
$\text{FF-GAN}^{\dag}$ & 94.6 & 92.5 & 89.7 & 85.2 & 77.2 & 61.2 \\
$\text{TP-GAN}^{\dag}$ & 98.7 & 98.1 & 95.4 & 87.7 & 77.4 & 64.6 \\
\midrule
$\text{LightCNN}^{*}$ & 98.6 & 97.4 & 92.1 & 62.1 & 24.2 & 5.5 \\
LB-GAN(Ours) & \textbf{99.1} & \textbf{98.9} & \textbf{96.7} & \textbf{91.0} & \textbf{80.3}& \textbf{65.4} \\
\bottomrule
\end{tabular}
}
\end{table}

\begin{table}[!tbp]
\small
\def\arraystretch{1.2}%  control the height
\centering
\caption{Performance comparisons on IJB-A. DA-GAN is trained on IJB-A with additional generated images, while our LB-GAN and FF-GAN are not trained on IJB-A. FF-GAN is the previous best frontalization method on IJB-A.}
\label{ijba}
\scalebox{0.9}
{
\begin{tabular}{ccccc}
\toprule
Method& \multicolumn{2}{c}{ACC(\%)} & \multicolumn{2}{c}{AUC(\%)} \\
\midrule
Metric((\%))& @FAR=0.01& FAR=0.001& Rank-1 & Rank-5 \\
\midrule
DR-GAN& 77.4$\pm$2.7& 53.9$\pm$4.3& 85.5$\pm$1.5 & 94.7$\pm$1.1 \\
FF-GAN& 85.2$\pm$1.0& 66.3$\pm$3.3& 90.2$\pm$0.6 & 95.4$\pm$0.5 \\
DA-GAN& 97.6$\pm$0.7& 93.0$\pm$0.5& 97.1$\pm$0.7 & 98.9$\pm$0.3 \\
\midrule
LB-GAN(Ours) & 92.3$\pm$0.8& 80.4$\pm$1.9& 94.7$\pm$0.6 & 97.0$\pm$0.4 \\
\bottomrule
\end{tabular}
}
\end{table}

\begin{table}[!tbp]
\small
\def\arraystretch{1.2}%  control the height
\centering
\caption{Mean head pose estimation errors (in degree) on Multi-PIE predicted by THPE.}
\label{thpe}
\begin{tabular}{cccccc}
\toprule
 & $\pm 30^{\circ}$& $\pm 22.5^{\circ}$ & $\pm 15^{\circ}$ & $\pm 7.5^{\circ}$ & $0^{\circ}$ \\
\midrule
Genuine data & 3.0 & - & 3.2 & - & 2.1  \\
Synthesized data & 4.6 & 5.7 & 4.0 & 5.1 & 2.9\\
\bottomrule
\end{tabular}
\end{table}

\begin{figure}[!tbp]
\centering 
\includegraphics[width=0.45\textwidth]
{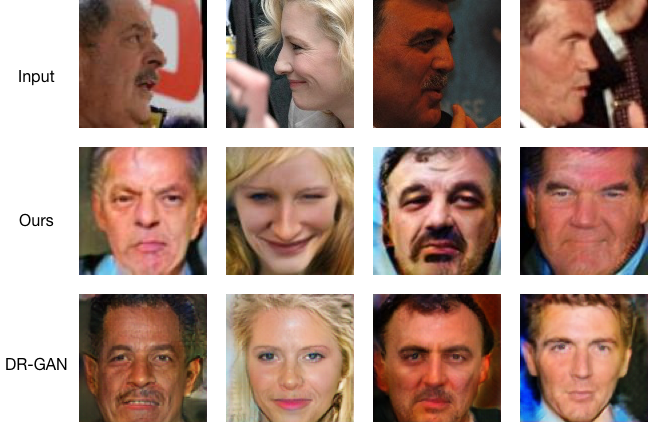}
\caption{Synthesis results on the IJB-A dataset.}
\label{IJBA}
\end{figure}

\textbf{Visual Quality.} The results in Fig. \ref{pose} show how our LB-GAN rotates the face images in Multi-PIE to specific poses. The poses are not limited to the 13 discrete values because we can produce any continuous value through interpolation \cite{radford2015unsupervised}. For instance, we average the remote codes for $15^{\circ}$ and $0^{\circ}$ to get the one for ${7.5^{\circ} }$. It will be hard for human observers to find the evidence of forgery on our results. Samples synthesized by interpolating identity representations \cite{DRGAN} are reported in Fig. \ref{ID}. Given two images of different subjects, we extract identity representations from the bottleneck layer of $G_E$ and then generate new representations through interpolation. Fed by those new representation, $G_E$ will synthesize new images with ``fused'' identities. We can see that the semantic changes in those images are very smooth and the visual realism is also very desirable. Frontalization results on IJB-A are shown in Fig. \ref{IJBA}. We compare with results produced by DR-GAN \cite{DRGAN}. The background of the input and the produced images looks different. We argue that for face frontalization in such challenging unconstrained environments, the concentration should be put on the facial parts. Despite that yaw angles are very large, our model still produces plausible results. DR-GAN produces comparable results, but the identities of some produced samples look very different from the original inputs.

\textbf{Pose-invariant Face Recognition.} To evaluate the capacity of identity preservation, we first use our model to frontalize profile face images in Multi-PIE and IJB-A and then evaluate face recognition performances through those produced images. We employ LightCNN \cite{he2017learning} as our feature extractor. For Multi-PIE, we make comparisons with DR-GAN, TP-GAN \cite{huang2017beyond} and FF-GAN \cite{yin2017towards}. The results are reported in Table \ref{multipie}. Note that TP-GAN and FF-GAN can only produce frontal view images. It can be observed that our results outperform them, especially for extreme poses, which indicates that our LB-GAN is able to preserve better identity information. For IJB-A, we make comparisons with DR-GAN, FF-GAN and DA-GAN \cite{zhao2017dual}. We report the results on Table \ref{ijba}. Note that DA-GAN is trained on IJB-A but FF-GAN and our LB-GAN are not. The effectiveness of our method is proved by the superior performance over FF-GAN.

\textbf{Head Pose Estimation.} To test whether our model is able to give correct responses to the remote code, we make head pose estimations on Multi-PIE. A \textbf{t}hird-party \textbf{h}ead \textbf{p}ose \textbf{e}stimator (THPE) \footnote{\url{https://github.com/guozhongluo/head-pose-estimation-and-face-landmark}} is employed. We simply call the high-level interface to train the model and then get the predicted yaw angles. The output of THPE is continuous angle value. Note that THPE is trained on the 300W dataset \cite{sagonas2013300} which consists only face images with yaw angles within $\pm {30^{\circ} }$. So only the images whose yaw angles are within this range are tested. The mean pose estimation errors are reported in Tabel \ref{thpe}. Those images with yaw angles of $\pm 7.5^\circ$ and $\pm 22.5^\circ$ are produced through interpolation. We can see that the mean errors made by THPE on the genuine and the synthesized data across all poses are very similar, which indicates that LB-GAN has the ability to control the pose of the output. The error of the images produced by interpolation is higher but still within an acceptable range.

\subsection{Ablation Study}

\begin{figure}[!tbp]
\begin{center}
\includegraphics[width=0.45\textwidth]{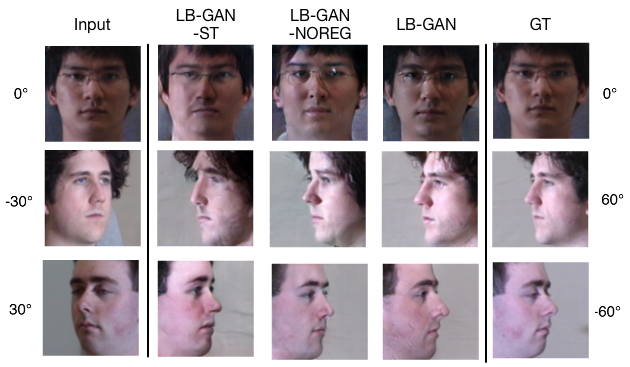}
\end{center}
\caption{Qualitative comparisons on synthesis results between LB-GAN and its variants. The degrees on the left side are the yaw angles of the inputs, and those on the right side are set by the remote codes.}
\label{ablation}
\end{figure}

\begin{table}[!tbp]
\small
\centering
\def\arraystretch{1.2}%  control the height
\caption{Identification rate (\%) comparison of model variations of our LB-GAN on Multi-PIE.}
\label{tab4}
\scalebox{0.8}
{
\begin{tabular}{ccccccc}
\toprule
Method& $\pm 15^{\circ}$& $\pm 30^{\circ}$ & $\pm 45^{\circ}$ & $\pm 60^{\circ}$ & $\pm 75^{\circ}$ & $\pm 90^{\circ}$\\
\midrule
LB-GAN & \textbf{99.1} & \textbf{98.9} & \textbf{96.7} & \textbf{91.0} & \textbf{80.3}& \textbf{65.4} \\
LB-GAN-ST & 93.7 & 92.1 & 85.3 & 83.2 & 72.0 & 59.9 \\
LB-GAN-NOREG & 96.8 & 93.9 & 91.2 & 87.6 & 77.6 & 63.5 \\
\bottomrule
\end{tabular}
}
\end{table}

\begin{table}[!tbp]
\small
\centering
\def\arraystretch{1.2}%  control the height
\caption{Performance comparison of model variations of our LB-GAN on IJB-A.}
\label{tab5}
\scalebox{0.8}
{
\begin{tabular}{ccccc}
\toprule
Method& \multicolumn{2}{c}{ACC(\%)} & \multicolumn{2}{c}{AUC(\%)} \\
\midrule
Metric((\%))& @FAR=0.01& FAR=0.001& Rank-1 & Rank-5 \\
\midrule
LB-GAN & \textbf{92.3$\pm$0.8}& \textbf{80.4$\pm$1.9}& \textbf{94.7$\pm$0.6} & \textbf{97.0$\pm$0.4} \\
LB-GAN-ST& 87.3$\pm$1.2& 70.2$\pm$2.2& 92.4$\pm$0.7 & 96.1$\pm$0.5 \\
LB-GAN-NOREG& 89.6$\pm$1.1& 74.6$\pm$2.0& 93.8$\pm$0.7 & 96.7$\pm$0.4 \\
\bottomrule
\end{tabular}
}
\end{table}

In this section, we demonstrate the effectiveness of our proposed two-stage training method and regularization items through an ablation study. Both qualitative and quantitative results are compared to make a comprehensive understanding. Specifically, we investigate the following model variations:

\begin{itemize}
   \item{LB-GAN-ST: All the components of the networks are trained jointly in a single stage through the conventional manner \cite{GAN}.}
   \item{LB-GAN-NOREG: The network is trained in the same way as LB-GAN. $L_{csc}$ is removed. $L{2^{\prime}}$ loss is replaced by $L2$ loss.}
\end{itemize}

A visual comparison is shown in Fig. \ref{ablation}. The face recognition performances on Multi-PIE and IJB-A are reported in Table \ref{tab4} and \ref{tab5} respectively. It can be observed that LB-GAN produces the most visually appealing results as well as achieve the best verification performance. The inferior performance of LB-GAN-ST indicates that two-stage training method is very important for our model. The effectiveness of regularization items is validated by the comparison between LB-GAN-NOREG and LB-GAN. As shown by the top row of images, the model tends to make superfluous manipulations and changes the contour of the input face obviously without those regularization items. Those observations prove that the regularization items can guide our model to put more attention on optimizing the facial part of its output. 

\section{Conclusion}
This paper has proposed LB-GAN for multi-view face image synthesis by decomposing the synthesis process into two subtasks. Input face images are first transformed into a frontal view by the face normalizer and then rotated to a specified angle by the face editor. A novel two-stage training method has also been accordingly proposed to help accomplish the two subtasks smoothly. To further improve the performance, conditional self-cycle loss and improved $L2$ loss have been integrated into LB-GAN. Experimental results have shown that our method is able to alter the pose of an input face image and keep the visual appearance photo-realistic simultaneously. Besides, our method obtains state-of-the-art face recognition results on publicly available datasets. In the future, we will investigate on the method to control more facial attributes, e.g., expression, race and gender.

% The file named.bst is a bibliography style file for BibTeX 0.99c
\bibliographystyle{named}
\bibliography{RF}

% \begin{thebibliography}{10}

% \bibitem{huang2017beyond}
% Huang, Rui and Zhang, Shu and Li, Tianyu and He, Ran.
% \newblock Beyond Face Rotation: Global and Local Perception GAN for Photorealistic and Identity Preserving Frontal View Synthesis.
% \newblock {\em arXiv preprint arXiv:1702.01983}, 2017.
% \end{thebibliography}

\end{document}